\documentclass[10pt,a4paper]{article}

\usepackage[T1]{fontenc}
\usepackage[utf8]{inputenc}
\usepackage{authblk} 
\usepackage[margin=1in]{geometry}
\usepackage{graphicx}
\usepackage{booktabs} 
\usepackage{amsmath,amssymb}
\usepackage{url}
\usepackage{xcolor}

\usepackage{microtype}
\usepackage{placeins} 
\usepackage{titlesec} 
\usepackage{multirow}
\usepackage{float}

\usepackage{hyperref}       

\usepackage{amsfonts}       
\usepackage{nicefrac}       
\usepackage{microtype}      
\usepackage{xcolor}         
\usepackage{amsthm} 
\usepackage{cleveref}
\usepackage{placeins}

\usepackage[utf8]{inputenc}
\usepackage[margin=1in]{geometry}
\usepackage{lmodern}
\usepackage{microtype}
\usepackage{amsmath,amssymb}
\usepackage{booktabs}
\usepackage{graphicx}
\usepackage[numbers,sort&compress]{natbib}
\newcommand{\extended}{\textsc{Ext}}
\newcommand{\normmodel}{\textsc{Norm}}
\newcommand{\rms}{\operatorname{RMS}}
\newcommand{\score}[2]{$#1\pm#2$}
\titleformat{\section}
  {\normalfont\Large\bfseries\raggedright}{\thesection.}{1em}{}
\titleformat{\subsection}
  {\normalfont\large\bfseries\raggedright}{\thesubsection.}{1em}{} 
\titleformat{\subsubsection}
  {\normalfont\normalsize\bfseries\raggedright}{\thesubsubsection.}{1em}{} 
\usepackage{hyperref}
\hypersetup{
    colorlinks=true,
    linkcolor=blue,
    filecolor=magenta,      
    urlcolor=blue,
    citecolor=blue,
    pdftitle={Do Language Models Need a Trainable Input Embedding Table? Fixed Minimal Token Codes at 1.7B-Class Scale},
    pdfauthor={Andrey Bochkov},
}

\title{
Do Language Models Need a Trainable Input Embedding Table?\\
Fixed Minimal Token Codes at 1.7B-Class Scale
}

\author[1]{Andrey Bochkov\thanks{Corresponding author: andrey.bochkov@gmail.com}}

\newcommand{\learned}{\textsc{Learned}}
\newcommand{\binary}{\textsc{Binary16}}
\newcommand{\gfmodel}{\textsc{GF2}}

\date{} 

\begin{document}
\maketitle

\begin{abstract}
A token must be identifiable before a language model can learn how to
use it. Must its input coordinates also be learned? We replace the
trainable input table of a decoder-only Transformer with a predetermined
16-bit identifier for each of 49,152 tokens. Repetition lifts the code
to width 2048 without a learned projection. We train five models from
scratch to 100B prediction targets each: a learned-input control and
four fixed-input variants using canonical codes, an invertible recoding,
an extra Transformer block, or row-normalized codes. At the endpoint,
the four fixed-input models attain 51.44--53.34\% HellaSwag normalized
accuracy, 70.46--71.49\% PIQA accuracy, and 42.29--44.52\% LAMBADA
accuracy. The normalized model attains the upper endpoint on all three;
the learned control scores 57.79\%, 72.69\%, and 47.72\%, respectively.
At unchanged depth and parameter count, the normalized model's LAMBADA
advantage over canonical codes is 1.80 percentage points at 48.44B
targets and 1.77 points at 100B. A complementary smaller-scale study
observes successful fixed-code learning across three training seeds per
interface. At matched backbone depth, the compact input buffer stores
128 times fewer scalar values than the untied learned input table.
The construction separates symbol identification from learned lexical
coordinates and provides a working model family in which to study how
shared computation builds representations from immutable inputs.
\end{abstract}

\section{Introduction}

An input embedding table plays two roles that are easy to conflate.
A token ID selects a particular row; training determines the coordinates
stored in that row. The first operation identifies the symbol. The second
allows its input representation to change independently of other input
rows. A learned embedding can carry semantic information. A different
question is whether semantic structure must already be present in
independently learned input coordinates for a model to use language
effectively.

We study this distinction by making token identity explicit and
immutable. Given a vocabulary of size $V$, its members can be
distinguished by $K=\lceil\log_2 V\rceil$ binary coordinates. For our
main vocabulary, $K=16$. We present these coordinates directly to a
Transformer by repeating them to the required input width. The model
has neither an independently trainable input vector per token nor a
dedicated learned embedding generator. Its contextual network and
untied output head remain trainable.

This construction gives an extreme but inspectable ablation. The
materialized input has 2048 coordinates, yet all its token-dependent
values are specified by a 16-bit identity. Two tokens can be
distinguished exactly without either receiving its own trainable input
parameters. What an injective code guarantees is distinguishability;
how well a finite network learns to use those distinctions is an
empirical question.

We address that question at two common training checkpoints, 48.44B and
100B processed prediction targets, for five 1.7--1.8B-class models.
A learned table and a canonical binary code provide the primary
comparison. Three additional configurations change the code assignment,
the scale of the fixed input, or the allocation of parameters to
contextual depth. The second checkpoint matters: an effect observed
partway through a cosine learning-rate schedule need not characterize
the trained endpoint.

The results distinguish three observations. First, the four fixed-input
configurations reach a similar absolute range on several informative
tasks, despite differences in assignment, normalization, and depth.
Second, the differences within that range are structured: the
normalized model has higher LAMBADA accuracy than the canonical model
at both checkpoints, while the relative ordering of the normalized and
deeper models on WikiText reverses. Third, learned inputs retain a
measurable advantage on several tasks. The evidence thus distinguishes
the \emph{utility} of independently adjustable input rows from their
\emph{necessity} for the capabilities observed here.

The interface has two further uses. For representation research, the
input associated with any token can be enumerated and held fixed
throughout training. For memory-constrained implementations, a compact
codebook replaces a much larger table; its storage cost can be computed
exactly. These uses do not require the fixed-input model to outperform
the learned control.

\section{Related work}

\paragraph{What learned embeddings establish.}
Trainable lexical representations are a standard component of neural
language models \citep{bengio2003neural,vaswani2017attention}.
Distributional methods demonstrate that useful lexical relationships
can be expressed through the geometry of learned vectors
\citep{mikolov2013efficient,pennington2014glove}. Our experiment asks
a complementary architectural question: can a contextual network
construct useful token-dependent features when the coordinates at its
input never move?

\paragraph{Freezing and sharing lexical parameters.}
\citet{kocmi2017exploration} study embedding initialization and frozen
embeddings, including sensitivity to scale. \citet{welch2020improving}
examine initialization, freezing, and input/output tying in
low-compute language modeling. \citet{hoffer2018fix} investigate fixed
classifiers and include an LSTM language-modeling experiment with fixed
lexical structure. \citet{bogoychev2021parameters} study freezing
Transformer components, including tied input/output embeddings.
Input/output treatment matters: we retain a full, trainable, untied
output projection. Weight tying \citep{press2017using} offers a different
way to avoid storing two independent vocabulary matrices.

\paragraph{Compact and generated representations.}
Factorization \citep{lan2020albert}, hash embeddings
\citep{svenstrup2017hash}, compositional codes
\citep{shu2018compositional}, and subword reconstruction
\citep{sasaki2021subword} replace independently learned full-width rows
with more structured parameterizations. ALONE generates word
representations from fixed word-specific filters and shared learned
components \citep{takase2020alone}. Our restriction is that both the
minimal identity code and its lift are predetermined: the first learned
computations belong to the standard backbone, not to an additional
lexical generator.

\paragraph{Fixed observations and tokenization.}
Frozen visual Unicode representations supply fixed observations of token
form \citep{bochkov2025emergent}. MaskBit uses bit-token inputs in image
generation \citep{weber2024maskbit}. Character-aware and byte-level
models change the units from which text is modeled
\citep{kim2016character,xue2022byt5}. In our controlled comparisons,
the tokenizer and the input sequence of token IDs are retained. It is
the map from each existing ID to the residual stream that changes.

\section{Token-identity interfaces}

\subsection{Identity width and computational width}

For token $t\in\{0,\ldots,V-1\}$, the learned control uses
$x(t)=E_{\mathrm{in}}[t]$, where
$E_{\mathrm{in}}\in\mathbb{R}^{V\times d}$. Our canonical code is the
little-endian binary expansion
$$
c(t)_j=\left\lfloor t/2^j\right\rfloor\bmod 2,
\qquad j=0,\ldots,K-1.
$$
There are $2^K$ bit strings of width $K$, so an injective
fixed-length binary code requires $2^K\geq V$. Canonical coding attains
the bound. Here $V=49{,}152$ and $K=16$. ``Minimal'' refers to this
binary identification problem, not to a universal lower bound on
real-valued embedding dimension.

For model width $d=sK$, repeat the code $s$ times:
$$
x(t)=Rc(t),\qquad
R=\begin{bmatrix}I_K\\ \vdots\\ I_K\end{bmatrix}
\in\mathbb{R}^{d\times K},\qquad R^\top R=sI_K.
$$
With $d=2048$, $s=128$. The lift preserves distinctness without adding
independent coordinates or trainable parameters. The raw input vectors
lie in a subspace of dimension at most 16. A linear map $W$ applied
directly to them acts as $WRc(t)$, a shared response to the code
coordinates. Nonlinear and contextual processing is not restricted
to the same input span.

\subsection{Assignment and scale}

\paragraph{Invertible recoding.}
The \gfmodel{} variant presents $R\widetilde c(t)$, where
$\widetilde c(t)=Ac(t)\bmod 2$ and $A\in\mathrm{GL}(16,2)$.
The fixed matrix, generated with code seed 12345, has binary rank 16
and minimum row and column Hamming weights of four. Its offset is zero.
This map preserves identity but can change Hamming distance, input norm,
and which tokens share active coordinates. The vocabulary occupies
only a subset of the full hypercube.

\paragraph{Row normalization.}
The \normmodel{} variant preserves the support and direction of each
nonzero canonical code but sets its RMS to the learned input's
initialization scale:
$$
\widehat c(t)=
\begin{cases}
0.02\, c(t)/\rms(c(t)), & c(t)\neq 0,\\
0, & c(t)=0,
\end{cases}
\qquad
\rms(c)=\sqrt{K^{-1}\sum_{j=0}^{K-1}c_j^2}.
$$
Its full-width input is $R\widehat c(t)$. The stored values are
real-valued and derived from the 16-bit identifier; they are not
quantized learned weights. Each nonzero row has full-width norm
$0.02\sqrt d$. Distinct code supports remain distinct, so normalization
preserves identity.

For comparison, the squared norm of an unnormalized lifted code of
Hamming weight $h$ is $128h$. A token with eight active bits thus enters
the residual stream with norm $32$, versus approximately $0.905$ after
row normalization. Pre-attention RMSNorm acts on the attention branch;
the raw vector still enters the residual addition. This gives a reason
to test scale even in a pre-normalized architecture. The intervention
changes both typical scale and between-token variation in norm.

\paragraph{Storage and output.}
The implementations store compact $V\times16$ buffers, retrieve a
row, and tile it. Normalization is computed once in FP32. GF2 also
stores its matrix and offset. The identity rules admit on-demand
generation; the evaluated implementations use compact lookup.
All models predict through a trainable untied
$W_{\mathrm{out}}\in\mathbb{R}^{V\times d}$.

\section{Experimental design}

\subsection{Models and comparisons}

Five decoder-only models are trained from scratch. The common backbone
uses width 2048, 32 attention heads, rotary positional encodings,
RMSNorm, and SwiGLU with intermediate width 8192. Context length is
2048; dropout and projection biases are disabled. Four configurations
have 24 blocks. \extended{} uses the canonical input and a 25th standard
block of the same width.

\begin{table}[t]
\centering
\small
\caption{Large-scale configurations. Every model retains 100,663,296
trainable output-head parameters. Fixed input entries are persistent
buffer values, not trainable parameters.}
\label{tab:params}
\begin{tabular}{lrrr}
\toprule
Configuration & Blocks & Trainable parameters & Fixed input entries\\
\midrule
\learned{}: untied learned table & 24 & 1,812,039,680 & 0\\
\binary{}: canonical codes & 24 & 1,711,376,384 & 786,432\\
\gfmodel{}: linear recoding & 24 & 1,711,376,384 & 786,704\\
\extended{}: canonical codes, extra block & 25 & 1,778,489,344 & 786,432\\
\normmodel{}: fixed row-normalized codes & 24 & 1,711,376,384 & 786,432\\
\bottomrule
\end{tabular}
\end{table}

The learned input table contains $Vd=100{,}663{,}296$ parameters,
5.56\% of the untied baseline. One block contains
$$
4d^2+3d\,d_{\mathrm{FFN}}+2d=67{,}112{,}960
$$
parameters. Adding one block therefore reallocates about two-thirds
of the removed capacity and leaves Ext 1.85\% below Learned in
trainable parameter count. It is an approximate parameter match using
whole standard blocks, not a compute match.

Binary16 is the common reference for three comparisons: GF2 changes
assignment, Norm changes row scaling at the same depth and parameter
count, and Ext changes depth at the same input rule. Learned supplies
both a backbone-matched reference for the 24-block models and an
approximately parameter-matched reference for Ext.

\subsection{Training data and optimization}

Large-scale pretraining uses the \texttt{fineweb-edu-dedup} component
of \texttt{HuggingFaceTB/smollm-corpus}, derived from FineWeb-Edu
\citep{penedo2024fineweb}. The preprocessing configuration covers 234
source shards. Training and validation documents are separated using a
seeded identifier hash, with a text-hash fallback and validation
threshold 0.001. The tokenizer comes from
\texttt{HuggingFaceTB/SmolLM2-1.7B}; no model weights are initialized
from SmolLM2.

Sampling stays within document boundaries. A document of at least 2049
tokens is selected uniformly among eligible documents in a resident
shard, then a valid window start is sampled uniformly. A window yields
2048 next-token targets. Processed-target counts include repeated or
overlapping windows and are not counts of unique corpus tokens.

AdamW uses peak learning rate $1.5\times10^{-4}$, minimum scheduled
rate $10^{-5}$, 2000 warmup updates, and cosine decay over 382,500
configured updates. Betas are $(0.9,0.95)$, epsilon is $10^{-8}$,
matrix weight decay is 0.01, and gradient clipping is 1.0. Parameters
remain FP32 with BF16 autocast and gradient checkpointing. The global
update contains 262,144 prediction targets.

All five trajectories are evaluated at the same two budgets:
\begin{center}
\begin{tabular}{lrr}
\toprule
Checkpoint & Completed updates & Processed prediction targets\\
\midrule
Intermediate (48.44B) & 184,800 & 48,444,211,200\\
Final (100B) & 381,470 & 100,000,071,680\\
\bottomrule
\end{tabular}
\end{center}
There is one large-scale trajectory per configuration, using training
seed 42. The two checkpoints of a trajectory are longitudinal
observations. The original three runs and the two additional runs
match the global batch and sampling recipe, rather than the exact
sequence of sampled windows (Appendix~\ref{app:training}).

\subsection{Smaller-scale replication}

A separate study uses 32-block models with width 1024, 32 heads,
context 1024, RoPE, GELU, an untied output head, and a 65,536-entry
tokenizer. Learned and fixed models contain approximately 537.2M and
470.1M trainable parameters. Training uses a FineWeb-Edu/Cosmopedia
mixture with sampling probabilities 0.8/0.2
\citep{penedo2024fineweb,benallal2024cosmopedia} and a common
document-level validation split. Learned, canonical, and affine-coded
inputs each have three training seeds. This series assesses repeated
training under a smaller recipe, separately from the five large
trajectories.

\subsection{Evaluation}

Base models are evaluated by likelihood with the Language Model
Evaluation Harness \citep{evalharness}. The suite includes HellaSwag,
ARC, PIQA, WinoGrande, OpenBookQA, CommonsenseQA, MMLU, LAMBADA, and
WikiText \citep{zellers2019hellaswag,clark2018arc,bisk2020piqa,
sakaguchi2020winogrande,mihaylov2018openbookqa,talmor2019commonsenseqa,
hendrycks2021mmlu,paperno2016lambada,merity2017pointer}. Multiple-choice
tasks use zero-shot evaluation; MMLU is also reported at five shots.
Raw accuracy and task-defined normalized accuracy are labeled separately.

All five intermediate evaluations and both final Ext/Norm evaluations
use Harness 0.4.10, Transformers 4.56.1, BF16 inference, batch size
one, context limit 2048, no automatic BOS insertion, no chat template,
and full selected splits. Final Learned, Binary16, and GF2 scores
come from the original endpoint evaluation. We identify the evaluation
batch in the full tables. In particular, direct WikiText comparisons
use the same documented scoring configuration
(Appendix~\ref{app:audit}). Harness standard errors describe evaluation
uncertainty, not training-seed variation.

\section{Results}

\subsection{Four fixed-input configurations at 100B}

\begin{table}[t]
\centering
\small
\caption{Final checkpoints at 100,000,071,680 prediction targets each.
Accuracies are percentages. ``Norm.'' in a metric name denotes the
benchmark scoring rule; the Norm column denotes the input-normalized
model. The full endpoint table with standard errors is in
Appendix~\ref{app:full}.}
\label{tab:final}
\begin{tabular}{lrrrrr}
\toprule
Metric & Learned & Binary16 & GF2 & Ext & Norm\\
\midrule
HellaSwag norm.\ accuracy & 57.79 & 52.40 & 51.44 & 52.85 & 53.34\\
ARC-Easy norm.\ accuracy & 66.04 & 61.53 & 60.73 & 63.47 & 64.39\\
ARC-Challenge norm.\ accuracy & 37.63 & 34.04 & 34.22 & 35.49 & 36.09\\
PIQA accuracy & 72.69 & 70.51 & 71.16 & 70.46 & 71.49\\
PIQA norm.\ accuracy & 72.14 & 71.11 & 72.14 & 71.71 & 72.09\\
WinoGrande accuracy & 58.56 & 55.33 & 55.01 & 54.70 & 57.46\\
OpenBookQA norm.\ accuracy & 37.80 & 38.00 & 36.80 & 36.00 & 35.00\\
LAMBADA accuracy & 47.72 & 42.75 & 42.29 & 42.85 & 44.52\\
LAMBADA perplexity $\downarrow$ & 12.88 & 17.91 & 18.47 & 17.60 & 16.53\\
\bottomrule
\end{tabular}
\end{table}

The four fixed-input models reach a narrow range on several tasks
where their accuracy is well above uniform choice
(Table~\ref{tab:final}). Across the four configurations, PIQA accuracy
spans 1.03 percentage points, from 70.46\% to 71.49\%; HellaSwag
normalized accuracy spans 1.90 points, from 51.44\% to 53.34\%.
On LAMBADA, which requires predicting the held-out last word rather
than selecting among supplied options, accuracy ranges from 42.29\%
to 44.52\%.

These ranges describe four different restrictions, not repeated
initializations of one model. Their interpretation is that useful
performance persists under a change of code assignment, a large change
of input scale, and the addition of a contextual block. Close results
on PIQA do not imply equally small changes everywhere: ARC-Easy
normalized accuracy spans 3.66 points, and the ordering of fixed
variants differs by task.

Norm has the highest fixed-input endpoint point estimates on HellaSwag
normalized accuracy, both reported ARC normalized accuracies, PIQA raw
accuracy, WinoGrande, and LAMBADA accuracy. Learned remains stronger
on those measures. The gap is task-dependent: for Norm it is 1.20
percentage points on PIQA accuracy and 4.45 points on HellaSwag
normalized accuracy. A fixed input thus retains a measurable level
of capability; the utility of a learned table is also measurable and
is not represented by one universal penalty.

\subsection{Normalization at matched depth and parameter count}

\begin{table}[t]
\centering
\small
\caption{Two checkpoints of the same training trajectories. Entries are
intermediate $\rightarrow$ final scores. WikiText is shown for Ext
and Norm, for which both checkpoints have the same documented
document-level evaluation. Lower perplexity is better.}
\label{tab:long}
\begin{tabular}{lrrrr}
\toprule
Metric & Learned & Binary16 & Ext & Norm\\
\midrule
HellaSwag acc\_norm
& $54.24\to57.79$ & $49.47\to52.40$ & $49.25\to52.85$ & $49.90\to53.34$\\
ARC-Easy acc\_norm
& $63.05\to66.04$ & $60.48\to61.53$ & $60.19\to63.47$ & $61.41\to64.39$\\
LAMBADA accuracy
& $43.35\to47.72$ & $40.07\to42.75$ & $39.72\to42.85$ & $41.88\to44.52$\\
LAMBADA PPL
& $15.94\to12.88$ & $22.32\to17.91$ & $21.53\to17.60$ & $19.90\to16.53$\\
WikiText word PPL
& --- & --- & $20.71\to18.69$ & $20.36\to20.17$\\
\bottomrule
\end{tabular}
\end{table}

Binary16 and Norm have the same canonical bit supports, backbone depth,
and trainable parameter count. Their LAMBADA results distinguish a
persistent difference from a transient intermediate ordering. At
48.44B, Norm scores 41.88\% versus 40.07\% for Binary16, a difference
of approximately 1.80 percentage points using unrounded scores.
At 100B the corresponding scores are 44.52\% and 42.75\%, a difference
of 1.77 points. Norm also has lower LAMBADA perplexity at both
checkpoints (Table~\ref{tab:long}).

Relative to Learned, the LAMBADA accuracy gap at the endpoint is
4.97 points for Binary16 and 3.20 for Norm. The absolute Norm--Binary16
difference is similar at the two checkpoints, but the fraction of the
learned-input gap it accounts for changes because Learned also improves.
These statements concern point estimates along the observed trajectories.

At the endpoint, Norm also scores 2.86 points above Binary16 on
ARC-Easy normalized accuracy and 2.13 points above it on WinoGrande.
The result is task-specific: Binary16 has higher OpenBookQA normalized
accuracy, 38.00\% versus 35.00\%, and higher MMLU point estimates.
There is no need to count wins across related metrics to identify
the clearest positive finding: a fixed change of input scale is
associated with better last-word prediction at both measured budgets.

\subsection{Reallocating parameters to one standard block}

At 48.44B, Ext and Binary16 have HellaSwag normalized accuracies of
49.25\% and 49.47\%, respectively. At 100B the ordering reverses,
52.85\% versus 52.40\%. The endpoint difference is larger on ARC:
Ext reaches 63.47\% versus 61.53\% on ARC-Easy and 35.49\% versus
34.04\% on ARC-Challenge, both normalized. On LAMBADA accuracy the
two finish within 0.10 percentage points of one another.

The endpoint thus gives a more informative account than the midpoint
alone. The deeper model gains on selected tasks, particularly ARC,
while leaving other scores close to the canonical 24-layer model.
It remains below Learned on HellaSwag, both ARC normalized accuracies,
and LAMBADA. Bringing the total trainable parameter count to within
1.85\% of Learned through this particular allocation is not, by itself,
a substitute for independent input rows on those tasks.

Ext and Norm also offer a direct comparison on document-level
WikiText. At the midpoint, word perplexities are 20.71 and 20.36,
respectively. At the endpoint they are 18.69 and 20.17: Ext improves
more over the second interval and finishes with lower corpus
perplexity, while Norm finishes with better LAMBADA accuracy and
perplexity. Both are likelihood-based evaluations, but WikiText
scores a corpus and LAMBADA scores a selected target continuation.
Their different orderings are part of the empirical result.

\subsection{Replication at smaller scale}

\begin{table}[t]
\centering
\small
\caption{Smaller-scale experiment, three training seeds per interface.
Perplexity is computed per run before averaging. Relative range is
$(\max-\min)/\mathrm{mean}$ across the three validation perplexities.}
\label{tab:small}
\begin{tabular}{lrrrr}
\toprule
Input & Tokens/seed (B) & Mean validation loss & Mean PPL & Relative range\\
\midrule
Learned table & 17.093 & 0.893 & 2.44 & 4.8\%\\
Canonical Binary16 & 17.093 & 0.859 & 2.36 & 4.6\%\\
Affine Binary16 & 16.276 & 0.871 & 2.39 & 4.5\%\\
\bottomrule
\end{tabular}
\end{table}

The smaller study contributes six fixed-input runs and three learned
controls. At the same budget of 17.093B targets per seed, the
canonical-code model has mean validation perplexity 2.36 versus 2.44
for Learned. The affine model reaches 2.39 with a shorter budget.
The code assignment is fixed across seeds of each interface;
initialization and data order vary. These observations establish
repeated successful training in that regime. They are not pooled with
the large-model task scores: the tokenizer, architecture, training
mixture, and evaluation distribution differ.

\paragraph{Where the benchmarks are informative.}
On PIQA and HellaSwag, the fixed-input endpoints substantially exceed
the uniform-choice references of 50\% and 25\%, respectively.
LAMBADA adds a prediction task without supplied choices. By contrast,
endpoint MMLU and CommonsenseQA scores lie near their uniform-choice
references for both learned and fixed inputs. Similar scores on a
near-chance task are weak evidence of retained capability. We report
the complete suite and base interpretation on what individual metrics
actually discriminate.

\section{What the fixed interface makes possible}

\subsection{A research and teaching example}

Consider a token whose canonical code has eight active bits. Before
the model sees a single training document, we can write down its exact
2048-dimensional input, compute its norm, and identify every other
token with which it shares any chosen bit. After 100B prediction
targets, those facts remain true of the input. What has changed is the
function that maps this token and its context to predictions.

This is the distinction that makes the construction useful for
teaching. A distributional embedding can encode learned similarities;
the experiment does not question that observation. It separates
\emph{where a symbol is identified} from \emph{where the model learns
to use that identification}. A learned input table is one location
for lexical adaptation, not a prerequisite for the capabilities
demonstrated here.

It is also a useful boundary for experiments. For instance, one can
measure the residual norm and attention or MLP update norm on the same
token in matched contexts over training. With a learned table, both
the token's starting vector and the subsequent computation move.
With a fixed table, the starting vector is known. This does not align
all hidden coordinates across models, but it removes a specific source
of change rather than leaving it to interpretation.

\subsection{Input storage under a fixed memory budget}
\label{sec:memory}

The deployed representation need not store the repeated vectors.
The evaluated canonical and normalized interfaces store $VK$ values
rather than $Vd$, an exact $d/K=128$-fold reduction in input tensor
entries. Table~\ref{tab:memory} gives the corresponding weight payloads
at two storage precisions. The second column is also an exact account
of the FP32 tensor payload in the exports, excluding file headers.

\begin{table}[t]
\centering
\small
\caption{Input tensor payload at equal storage precision
($1$ MiB $=2^{20}$ bytes). These are input-side weight and buffer
sizes, not device peak-memory measurements.}
\label{tab:memory}
\begin{tabular}{lrr}
\toprule
Input representation & FP32 & BF16\\
\midrule
Untied learned table, $49{,}152\times2048$ & 384 MiB & 192 MiB\\
Compact Binary16 or Norm buffer, $49{,}152\times16$ & 3 MiB & 1.5 MiB\\
Difference at the same backbone depth & 381 MiB & 190.5 MiB\\
\bottomrule
\end{tabular}
\end{table}

For a memory-constrained runtime, including an on-device runtime, this
is a concrete design choice: store independent lexical weights, store a
compact buffer, or generate the canonical bits from token IDs. The
last option follows directly from the coding rule; the benchmarked
models used the buffer implementation. GF2 adds only 272 matrix and
offset entries to the compact buffer. Normalized codes can also be
regenerated from the ID and its Hamming weight.

The $128$-fold factor applies to the \emph{input tensor}, not the whole
model. The output projection and backbone remain, and Ext spends part
of the released weight budget on depth. With tied input/output weights,
the output matrix already provides the input table's storage, so the
same savings do not follow. The appropriate device experiment is to
measure model residency, peak working memory, and latency at a common
precision on actual hardware. The present storage accounting identifies
what such an experiment can save before runtime overhead is considered.

\subsection{Experiments enabled by the construction}

\paragraph{Geometry at fixed storage.}
A frozen Gaussian $V\times16$ codebook with identical row RMS and
the same tiled lift would match the compact buffer size, depth, and
number of trainable parameters of Norm. Comparing the two would
test binary support geometry against continuous random directions.
A full-width frozen Gaussian table asks a different question: it
preserves frozen coordinates but relaxes the rank and storage
restrictions. The binary construction's demonstrated distinction is
its minimal, deterministic specification; an accuracy advantage over
Gaussian inputs is an open empirical comparison.

\paragraph{Adaptive shared projections.}
For a learned projection $P$ acting on a fixed code $c$, an example
with upstream gradient $g$ contributes $gc^\top$ to the gradient of
$P$. Under zero/one coding, two token types therefore update overlapping
projection columns when their codes share active bits. Comparing
several injective assignments at fixed norm, projection size, and data
budget would test the role of this sharing. Learning $P$ and learning
a separate 16-dimensional row for each token are distinct experiments.

\paragraph{Reusing the identity rule.}
A common tokenizer and ID specification make the fixed interface
reproducible across contextual architectures. A separate attention-free
causal multi-mesh implementation uses Binary16 coordinates together
with glyph/form observations and structured fields
\citep{bochkov2026parametermonolith}. It exemplifies reuse of the
identity rule rather than a Binary16-only comparison. Likewise, an
unused binary code can identify a new vocabulary entry without
allocating an independently learned input row. Learning the new token's
use, and extending the output head if it must be predicted, are
separate operations.

\section{Experimental scope}

The five large configurations are individual trained trajectories.
Differences and ranges refer to their observed checkpoints. The smaller
series measures variation across three seeds per interface in its own
regime. Equal token budgets do not imply equal FLOPs across depths,
and task-level standard errors do not measure training-seed variance.
The architecture and accounting are explicit so that comparisons can
be extended along each of these dimensions independently.

\section{Conclusion}

All four fixed-input configurations complete 100B-target training.
A 16-bit predetermined identifier, presented through a fixed lift,
supports last-word prediction, commonsense completion, and corpus
language modeling in a wide decoder without a trainable input table.
Recoding, normalization, and an additional block preserve that broad
capability while producing distinguishable task profiles. In particular,
row normalization yields a similar LAMBADA accuracy advantage over
canonical input at both measured budgets, while extra depth gains on
selected endpoint tasks and yields the better final WikiText score of
the two additional configurations.

The result is both architectural and practical: symbol identity can
be held immutable while useful representations are learned downstream,
and the corresponding input tensor can be specified with 128 times
fewer stored values than an untied full-width table. This makes the
interface a compact implementation choice and a transparent starting
point for experiments on lexical representation.

\section*{Reproducibility and availability}

The release location is \texttt{https://huggingface.co/Bochkov}, with repositories for the three evaluated interfaces:

\begin{itemize}
    \item \texttt{https://huggingface.co/Bochkov/ab\_ext\_learned};
    \item \texttt{https://huggingface.co/Bochkov/ab\_ext\_binary16};
    \item \texttt{https://huggingface.co/Bochkov/ab\_ext\_gf2}.
\end{itemize}

Appendix~\ref{app:audit} identifies the additional checkpoints by their
training counters and evaluation metadata. The original training
implementation accepts externally shifted targets; the Hugging Face
evaluation adapter uses the standard internally shifted label convention.
Likelihood evaluation consumes logits.

\section*{Broader impact}

The intended uses are experimental language modeling, representation
analysis, and teaching. A fixed input specification changes how tokens
are represented, not the reliability of model outputs. The checkpoints
are base research models, not validated assistants for high-stakes use.

\bibliographystyle{plainnat}
\bibliography{main}

\clearpage
\appendix

\section{Complete internal evaluations}
\label{app:full}

The two tables below report all internal models at both budgets.
Accuracy values are percentages and their errors are in percentage
points. Perplexities and bits per byte use their native units.
The $\pm$ entries are Harness evaluation standard errors. No standard
error was reported for WikiText. Accuracy and normalized accuracy on
one task are alternative scoring rules, not independent replications.

\begin{table}[htbp]
\centering
\small
\caption{Final evaluations: 381,470 updates and 100,000,071,680
prediction targets per model. Learned, Binary16, and GF2 are from
the original endpoint batch; Ext and Norm are from the additional
endpoint batch. The dagger marks WikiText rows for which comparisons
across these two batches require reconciliation of the scoring records.}
\label{tab:full100}
\resizebox{\linewidth}{!}{%
\begin{tabular}{lrrrrr}
\toprule
Metric & Learned & Binary16 & GF2 & Ext & Norm\\
\midrule
HellaSwag acc
& \score{44.21}{0.50} & \score{40.70}{0.49} & \score{40.24}{0.49}
& \score{40.78}{0.49} & \score{41.09}{0.49}\\
HellaSwag acc\_norm
& \score{57.79}{0.49} & \score{52.40}{0.50} & \score{51.44}{0.50}
& \score{52.85}{0.50} & \score{53.34}{0.50}\\
ARC-Easy acc
& \score{71.63}{0.92} & \score{67.59}{0.96} & \score{66.84}{0.97}
& \score{67.55}{0.96} & \score{69.15}{0.95}\\
ARC-Easy acc\_norm
& \score{66.04}{0.97} & \score{61.53}{1.00} & \score{60.73}{1.00}
& \score{63.47}{0.99} & \score{64.39}{0.98}\\
ARC-Challenge acc
& \score{35.92}{1.40} & \score{32.34}{1.37} & \score{30.55}{1.35}
& \score{32.76}{1.37} & \score{33.53}{1.38}\\
ARC-Challenge acc\_norm
& \score{37.63}{1.42} & \score{34.04}{1.38} & \score{34.22}{1.39}
& \score{35.49}{1.40} & \score{36.09}{1.40}\\
PIQA acc
& \score{72.69}{1.04} & \score{70.51}{1.06} & \score{71.16}{1.06}
& \score{70.46}{1.06} & \score{71.49}{1.05}\\
PIQA acc\_norm
& \score{72.14}{1.05} & \score{71.11}{1.06} & \score{72.14}{1.05}
& \score{71.71}{1.05} & \score{72.09}{1.05}\\
WinoGrande acc
& \score{58.56}{1.38} & \score{55.33}{1.40} & \score{55.01}{1.40}
& \score{54.70}{1.40} & \score{57.46}{1.39}\\
OpenBookQA acc
& \score{27.60}{2.00} & \score{28.20}{2.01} & \score{25.20}{1.94}
& \score{26.00}{1.96} & \score{25.80}{1.96}\\
OpenBookQA acc\_norm
& \score{37.80}{2.17} & \score{38.00}{2.17} & \score{36.80}{2.16}
& \score{36.00}{2.15} & \score{35.00}{2.14}\\
CommonsenseQA acc
& \score{19.82}{1.14} & \score{20.56}{1.16} & \score{19.74}{1.14}
& \score{18.43}{1.11} & \score{21.13}{1.17}\\
MMLU 0-shot
& \score{25.32}{0.37} & \score{25.88}{0.37} & \score{26.11}{0.37}
& \score{25.13}{0.37} & \score{24.74}{0.36}\\
MMLU 5-shot
& \score{25.48}{0.37} & \score{25.57}{0.37} & \score{24.66}{0.36}
& \score{25.35}{0.37} & \score{25.04}{0.37}\\
LAMBADA accuracy
& \score{47.72}{0.70} & \score{42.75}{0.69} & \score{42.29}{0.69}
& \score{42.85}{0.69} & \score{44.52}{0.69}\\
LAMBADA perplexity
& \score{12.88}{0.42} & \score{17.91}{0.62} & \score{18.47}{0.63}
& \score{17.60}{0.61} & \score{16.53}{0.56}\\
WikiText word PPL$^\dagger$ & 16.50 & 18.58 & 19.03 & 18.69 & 20.17\\
WikiText byte PPL$^\dagger$ & 1.69 & 1.73 & 1.73 & 1.73 & 1.75\\
WikiText bits/byte$^\dagger$ & 0.76 & 0.79 & 0.79 & 0.7899 & 0.8106\\
\bottomrule
\end{tabular}}
\end{table}

\begin{table}[htbp]
\centering
\small
\caption{Intermediate evaluations: 184,800 updates and 48,444,211,200
prediction targets per model. All five columns use the same specified
Harness configuration, including the same document-level WikiText task.}
\label{tab:full48}
\resizebox{\linewidth}{!}{%
\begin{tabular}{lrrrrr}
\toprule
Metric & Learned & Binary16 & GF2 & Ext & Norm\\
\midrule
HellaSwag acc
& \score{41.84}{0.49} & \score{39.17}{0.49} & \score{38.60}{0.49}
& \score{38.79}{0.49} & \score{39.26}{0.49}\\
HellaSwag acc\_norm
& \score{54.24}{0.50} & \score{49.47}{0.50} & \score{48.39}{0.50}
& \score{49.25}{0.50} & \score{49.90}{0.50}\\
ARC-Easy acc
& \score{69.02}{0.95} & \score{65.32}{0.98} & \score{65.11}{0.98}
& \score{65.15}{0.98} & \score{65.74}{0.97}\\
ARC-Easy acc\_norm
& \score{63.05}{0.99} & \score{60.48}{1.00} & \score{60.65}{1.00}
& \score{60.19}{1.00} & \score{61.41}{1.00}\\
ARC-Challenge acc
& \score{34.90}{1.39} & \score{30.80}{1.35} & \score{32.08}{1.36}
& \score{29.52}{1.33} & \score{31.66}{1.36}\\
ARC-Challenge acc\_norm
& \score{37.03}{1.41} & \score{35.07}{1.39} & \score{33.19}{1.38}
& \score{32.42}{1.37} & \score{34.22}{1.39}\\
PIQA acc
& \score{71.60}{1.05} & \score{69.53}{1.07} & \score{69.21}{1.08}
& \score{70.08}{1.07} & \score{70.29}{1.07}\\
PIQA acc\_norm
& \score{71.27}{1.06} & \score{70.18}{1.07} & \score{69.42}{1.07}
& \score{69.70}{1.07} & \score{71.44}{1.05}\\
WinoGrande acc
& \score{55.88}{1.40} & \score{54.93}{1.40} & \score{51.85}{1.40}
& \score{53.67}{1.40} & \score{54.22}{1.40}\\
OpenBookQA acc
& \score{25.60}{1.95} & \score{26.40}{1.97} & \score{26.20}{1.97}
& \score{25.60}{1.95} & \score{25.40}{1.95}\\
OpenBookQA acc\_norm
& \score{35.60}{2.14} & \score{35.80}{2.15} & \score{36.80}{2.16}
& \score{37.60}{2.17} & \score{36.00}{2.15}\\
CommonsenseQA acc
& \score{19.25}{1.13} & \score{19.25}{1.13} & \score{19.00}{1.12}
& \score{20.64}{1.16} & \score{20.07}{1.15}\\
MMLU 0-shot
& \score{25.16}{0.37} & \score{25.79}{0.37} & \score{26.14}{0.37}
& \score{26.11}{0.37} & \score{24.92}{0.36}\\
MMLU 5-shot
& \score{25.59}{0.37} & \score{27.12}{0.37} & \score{25.97}{0.37}
& \score{26.21}{0.37} & \score{24.98}{0.36}\\
LAMBADA accuracy
& \score{43.35}{0.69} & \score{40.07}{0.68} & \score{38.50}{0.68}
& \score{39.72}{0.68} & \score{41.88}{0.69}\\
LAMBADA perplexity
& \score{15.94}{0.53} & \score{22.32}{0.81} & \score{23.45}{0.84}
& \score{21.53}{0.78} & \score{19.90}{0.71}\\
WikiText word PPL & 18.48 & 20.48 & 21.28 & 20.71 & 20.36\\
WikiText byte PPL & 1.73 & 1.76 & 1.77 & 1.76 & 1.76\\
WikiText bits/byte & 0.7869 & 0.8146 & 0.8249 & 0.8177 & 0.8130\\
\bottomrule
\end{tabular}}
\end{table}

\clearpage
\section{External quality references}

Table~\ref{tab:external} retains the locally evaluated SmolLM2 base
checkpoints from the original endpoint evaluation. Their reported
training budgets are approximately 2T, 4T, and 11T tokens for the 135M,
360M, and 1.7B models, respectively \citep{benallal2025smollm2}.
Their size, architecture, data, and optimization differ from ours.
They contextualize benchmark scores rather than control for the
lexical interface or establish a sample-efficiency ratio.

\begin{table}[htbp]
\centering
\small
\caption{External base checkpoints from the original evaluation batch.
Units follow Appendix~\ref{app:full}.}
\label{tab:external}
\begin{tabular}{lrrr}
\toprule
Metric & SmolLM2-135M & SmolLM2-360M & SmolLM2-1.7B\\
\midrule
HellaSwag acc & \score{35.36}{0.48} & \score{43.05}{0.49} & \score{53.38}{0.50}\\
HellaSwag acc\_norm & \score{43.02}{0.49} & \score{56.28}{0.50} & \score{71.43}{0.45}\\
ARC-Easy acc & \score{64.44}{0.98} & \score{70.24}{0.94} & \score{77.86}{0.85}\\
ARC-Easy acc\_norm & \score{58.75}{1.01} & \score{68.18}{0.96} & \score{73.36}{0.91}\\
ARC-Challenge acc & \score{28.07}{1.31} & \score{36.26}{1.40} & \score{44.37}{1.45}\\
ARC-Challenge acc\_norm & \score{29.61}{1.33} & \score{38.05}{1.42} & \score{47.27}{1.46}\\
PIQA acc & \score{68.44}{1.08} & \score{71.38}{1.05} & \score{76.99}{0.98}\\
PIQA acc\_norm & \score{68.39}{1.08} & \score{71.82}{1.05} & \score{77.20}{0.98}\\
WinoGrande acc & \score{52.57}{1.40} & \score{59.35}{1.38} & \score{65.98}{1.33}\\
OpenBookQA acc & \score{22.00}{1.85} & \score{24.80}{1.93} & \score{32.20}{2.09}\\
OpenBookQA acc\_norm & \score{32.60}{2.10} & \score{37.80}{2.17} & \score{44.40}{2.22}\\
CommonsenseQA acc & \score{19.90}{1.14} & \score{21.05}{1.17} & \score{41.69}{1.41}\\
MMLU 0-shot & \score{24.25}{0.36} & \score{25.47}{0.37} & \score{48.40}{0.41}\\
MMLU 5-shot & \score{25.15}{0.36} & \score{25.03}{0.37} & \score{50.06}{0.41}\\
LAMBADA accuracy & \score{42.97}{0.69} & \score{53.31}{0.70} & \score{67.51}{0.65}\\
LAMBADA perplexity & \score{19.06}{0.63} & \score{9.38}{0.27} & \score{4.44}{0.10}\\
WikiText word PPL & 23.14 & 17.12 & 11.62\\
WikiText byte PPL & 1.80 & 1.70 & 1.58\\
WikiText bits/byte & 0.85 & 0.77 & 0.66\\
\bottomrule
\end{tabular}
\end{table}

\section{Training and implementation records}
\label{app:training}

\paragraph{Large runs.}
Learned, Binary16, and GF2 were launched with two GPUs, eight sequences
per GPU per microbatch, and eight accumulation steps. Ext and Norm
were launched with four GPUs, eight sequences per GPU, and four
accumulation steps. With context 2048, both correspond to 262,144
prediction targets per update. Sampling is rank-specific; the launch
configuration uses 1024 microbatches per resident-shard visit. The
original trainer does not fully restore the sampler cursor and dedicated
sampling generator on resume, whereas the Ext/Norm trainer saves them.
Matching counters therefore specifies training exposure, not
sample-by-sample identity.

Linear layers and learned input weights use initialization standard
deviation 0.02. Attention output and FFN down projections use
$0.02/\sqrt{2L}$ for depth $L$. Ext follows this same rule at $L=25$.
RoPE base is 10,000 and RMSNorm epsilon is $10^{-5}$. The optimizer
applies no weight decay to one-dimensional parameters.

\paragraph{Counter provenance.}
For all five intermediate checkpoints, exported metadata record step
184,800 and 48,444,211,200 targets. For the three original final runs,
terminal logs record optimizer step 381,470 and a stopping counter of
100,000,071,680 targets. For Ext and Norm, final snapshot metadata
directly record both numbers. The small overshoot of the nominal
100B target is due to stopping at an optimizer-update boundary.

\paragraph{Smaller runs.}
The smaller series uses FP32 training, AdamW with peak learning rate
$4\times10^{-4}$, cosine decay with 150 warmup updates, weight decay
0.1, and clipping 1.0. The global microbatch is 16 sequences with
accumulation 200 and context 1024. The canonical implementation stores
a compact frozen table; the affine variant generates codes from fixed
buffers. The historical canonical audit includes zeroing the padding
row: if the padding ID is nonzero, the injectivity statement applies
to active non-padding tokens rather than to every stored row. The
large-scale canonical implementation uses the unmodified rule.
The smaller validation means and relative seed ranges are reported
from the three-run summaries and are not used to estimate uncertainty
in the large-scale benchmark differences.

\paragraph{Boundary case.}
In the large models, canonical token zero maps to zero. The zero-offset
GF2 map and row normalization preserve this code. In the bias-free
architecture an all-zero-code context produces zero hidden states and
uniform next-token probabilities, assuming finite weights. A zero-code
token after a nonzero context can receive information through causal
attention. The evaluation tokenizer identifies zero as EOS, BOS, and
padding; disabling automatic BOS insertion is therefore an important
part of the protocol.

\section{Evaluation and artifact records}
\label{app:audit}

\paragraph{Common intermediate protocol and additional endpoints.}
The Harness task identifiers are \texttt{hellaswag}, \texttt{arc\_easy},
\texttt{arc\_challenge}, \texttt{piqa}, \texttt{winogrande},
\texttt{openbookqa}, \texttt{commonsense\_qa}, \texttt{mmlu},
\texttt{lambada\_openai}, and \texttt{wikitext}. The evaluation seed
tuple is \texttt{0,1234,1234,1234}. The full selected splits contain
10,042 HellaSwag examples, 2,376 ARC-Easy, 1,172 ARC-Challenge,
1,838 PIQA, 1,267 WinoGrande, 500 OpenBookQA, 1,221 CommonsenseQA,
and 5,153 LAMBADA examples. MMLU uses the selected subject test sets at
zero and five shots. Ext and Norm final evaluations report the same
settings and coverage as their intermediate evaluations.

The WikiText task in this protocol is
\texttt{EleutherAI/wikitext\_document\_level},
\texttt{wikitext-2-raw-v1}, test split, with 62 documents. The
original endpoint audit instead records 173 rolling request units.
Documents and backend requests need not be the same unit, and the
available records do not establish their correspondence. Consequently,
the five-way intermediate WikiText table and the Ext--Norm
longitudinal comparison each have a common documented basis. The
original three endpoint WikiText scores are retained separately in
interpretation, even when displayed in the same table.

\paragraph{Original endpoint audit.}
The original evaluation of Learned, Binary16, GF2, and the three
external references reports identical sample/prompt multisets within
each task group. It records 68,671 multiple-choice requests, 56,168
MMLU candidate requests per shot setting, 5,153 LAMBADA requests, and
173 WikiText rolling requests per model. For five-shot MMLU, 1,508 of
56,168 candidate requests were marked as truncated. This count belongs
to that evaluation batch, not to the additional endpoint evaluations,
for which the summaries do not report request-level truncation counts.

\paragraph{Exports.}
Intermediate Learned, Binary16, and GF2 exports used strict
state-dictionary loading. Their input checks verify the learned table
shape, canonical coding, and the rank-16 zero-offset GF2 transformation,
respectively. On the intermediate Ext and Norm snapshots, training and
exported BF16 logits agreed on 40 short unpadded prompts per model;
the loss check verified internal HF label shifting. These verification
records concern those tested snapshots. Final Ext and Norm scores are
taken from the completed full-split evaluation outputs.

\paragraph{Reading the tables.}
LAMBADA perplexity concerns the target continuation. WikiText word and
byte perplexities normalize corpus negative log-likelihood by different
external units; bits per byte is the base-two logarithm of byte
perplexity. These are not three independent confirmations of the same
effect. All differences reported in the text are descriptive
checkpoint contrasts. In particular, a range over four configurations
does not estimate seed variance, and marginal Harness errors are not
paired tests of differences between models.

\end{document}